\documentclass{article}

\PassOptionsToPackage{hyphens}{url}

\usepackage{longcat_style,times}
\usepackage[T1]{fontenc}
\usepackage[utf8]{inputenc}
\usepackage{microtype}

\usepackage{amsmath,amssymb}
\usepackage{booktabs,multirow}
\usepackage[table,dvipsnames]{xcolor}
\usepackage{pgfplots}
\pgfplotsset{compat=1.18}

\usepackage{natbib}
\usepackage{xurl}
\usepackage{hyperref}

\definecolor{linkblue}{rgb}{0,0,0.5}
\hypersetup{
  colorlinks=true,
  citecolor=linkblue,
  linkcolor=linkblue,
  urlcolor=linkblue,
  pdftitle={Retrieval Grounding Latent Reasoning for Dense Retrieval},
  pdfauthor={Gang Zhou, Xiongxi Yu, Hu Tian, Yang Wei, Lu Pan, Ke Zeng, Shibiao Xu, Xiaolong Zheng}
}
\AtBeginDocument{\let\cite\citep}

\newcommand{\method}{\textsc{RGLT}}
\newcommand{\anchor}{\texttt{<anchor>}}

\newcommand{\emb}{\texttt{<|embed\_token|>}}

\newcommand{\shortpapertitle}{RGLT}
\renewcommand{\shorttitle}{\shortpapertitle}
\renewcommand{\headeright}{%
  \raisebox{-0.2\height}{\includegraphics[height=2em]{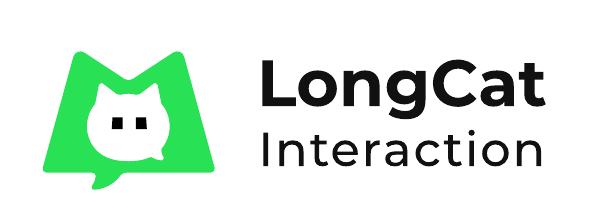}}%
}

\title{Retrieval Grounding Latent Reasoning for Dense Retrieval}
\author{%
  \href{https://openreview.net/profile?id=\%7EGang\%5FZhou6}{\textbf{Gang Zhou}}\textsuperscript{1,*,$\ddagger$}
  \quad
  \href{https://openreview.net/profile?id=\%7EXiongxi\%5FYu1}{\textbf{Xiongxi Yu}}\textsuperscript{2,*}
  \quad
  \href{https://openreview.net/profile?id=\%7EHu\%5FTian4}{\textbf{Hu Tian}}\textsuperscript{3,$\dagger$}
  \quad
  \href{https://openreview.net/profile?id=\%7EYang\%5FWei13}{\textbf{Yang Wei}}\textsuperscript{2,$\dagger$}
  \\[0.5ex]
  \href{https://openreview.net/profile?id=\%7ELu\%5FPan3}{\textbf{Lu Pan}}\textsuperscript{2}
  \quad
  \href{https://openreview.net/profile?id=\%7EKe\%5FZeng1}{\textbf{Ke Zeng}}\textsuperscript{2}
  \quad
  \href{https://openreview.net/profile?id=\%7EShibiao\%5FXu1}{\textbf{Shibiao Xu}}\textsuperscript{1,\S}
  \quad
  \href{https://openreview.net/profile?id=\%7EXiaolong\%5FZheng4}{\textbf{Xiaolong Zheng}}\textsuperscript{4,\S}
  \\[1ex]
  \normalfont\footnotesize
  \textsuperscript{1}School of Artificial Intelligence, Beijing University of Posts and Telecommunications
  \\[-0.1ex]
  \normalfont\footnotesize
  \textsuperscript{2}Meituan LongCat Interaction Team
  \\[-0.1ex]
  \normalfont\footnotesize
  \textsuperscript{3}School of Management Science and Engineering, Central University of Finance and Economics
  \\[-0.1ex]
  \normalfont\footnotesize
  \textsuperscript{4}Institute of Automation, Chinese Academy of Sciences
  \\[0.6ex]
  \normalfont\scriptsize
  \href{mailto:zhougang2023@bupt.edu.cn}{\texttt{zhougang2023@bupt.edu.cn}}
  \quad
  \href{mailto:yuxiongxi@meituan.com}{\texttt{yuxiongxi@meituan.com}}
  \quad
  \href{mailto:tianhu01@foxmail.com}{\texttt{tianhu01@foxmail.com}}
  \quad
  \href{mailto:weiyang14@meituan.com}{\texttt{weiyang14@meituan.com}}
  \\[-0.1ex]
  \normalfont\scriptsize
  \href{mailto:panlu02@meituan.com}{\texttt{panlu02@meituan.com}}
  \quad
  \href{mailto:zengke02@meituan.com}{\texttt{zengke02@meituan.com}}
  \quad
  \href{mailto:shibiaoxu@bupt.edu.cn}{\texttt{shibiaoxu@bupt.edu.cn}}
  \quad
  \href{mailto:xiaolong.zheng@ia.ac.cn}{\texttt{xiaolong.zheng@ia.ac.cn}}
}

\newcommand{\printauthornotes}{%
  \begingroup
    \renewcommand{\thefootnote}{\fnsymbol{footnote}}%
    \footnotetext[1]{Gang Zhou and Xiongxi Yu contributed equally to this work.}%
    \footnotetext[2]{Hu Tian and Yang Wei are corresponding authors.}%
    \footnotetext[3]{This work was conducted during Gang Zhou's internship at Meituan.}%
    \footnotetext[4]{Shibiao Xu and Xiaolong Zheng, as Gang Zhou's advisors, primarily provided guidance on topic selection for this work.}%
  \endgroup
}

\begin{document}

\maketitle
\printauthornotes

\bibliographystyle{iclr2026_conference}

\begin{abstract}
Reasoning-intensive retrieval requires text representations to capture not only semantic similarity, but also the reasoning needed to determine relevance under a given retrieval instruction. Existing reasoning-enhanced embedding models improve retrieval by incorporating reasoning information into dense representations, yet their supervision is typically dominated by the final retrieval objective. As a result, latent reasoning trajectories may learn shortcut reasoning patterns that preserve retrieval performance without producing meaningful incremental retrieval gains. We propose Retrieval Grounding Latent Reasoning (RGLT), a latent reasoning framework for dense retrieval that explicitly connects intermediate latent transitions with retrieval improvements. RGLT performs non-autoregressive reasoning in hidden space through an instruction-conditioned latent reasoning trajectory constructed from silent tokens. It combines process-supervised explicit-to-implicit distillation with retrieval-grounded supervision, using stage-wise CoT reconstruction to shape intermediate latent states and retrieval-effect credit to optimize incremental retrieval gains across the latent reasoning trajectories. Experiments on reasoning-intensive retrieval benchmarks show that RGLT consistently outperforms strong baselines while preserving efficient embedding inference.

\end{abstract}


\section{Introduction}

Reasoning-intensive retrieval requires text representations to capture not only semantic similarity, but also the multi-stage reasoning needed to identify relevant evidence. For specialized domains like mathematics, science and programming, target documents may share little surface-level similarity with the query. Their relevance may emerge only after the implicit constraints and intermediate concepts are resolved. Recent reasoning-intensive benchmarks show that dense retrievers effective on conventional semantic retrieval may struggle on retrieval tasks requiring multi-stage reasoning, while explicit reasoning methods exhibit substantially greater potential~\citep{bright}. These findings suggest that, in reasoning-intensive retrieval tasks, query representations should incorporate inferred evidence beyond literal semantics to identify relevant documents.

One line of research addresses this problem by expanding the query with a generated explicit Chain-of-Thought (CoT), and then encoding the enriched text for retrieval. Despite their effectiveness, these methods introduce autoregressive decoding latency and make retrieval quality highly dependent on the verbosity and lexical form of the generated rationale. Subsequent studies on continuous reasoning have explored replacing textual reasoning steps with hidden states, showing that multi-stage reasoning can work without verbalizing every intermediate stage in natural language~\citep{coconut,codi}. More recently, embedding models have further internalized explicit CoT into latent tokens to learn reasoning-enhanced representations, allowing the final embeddings to capture reasoning information without long autoregressive generation~\citep{laser,xetrieval}. These works establish latent reasoning as a promising direction for combining reasoning ability with efficient dense retrieval.

\begin{figure}[!t]
\centering
\includegraphics[width=\columnwidth]{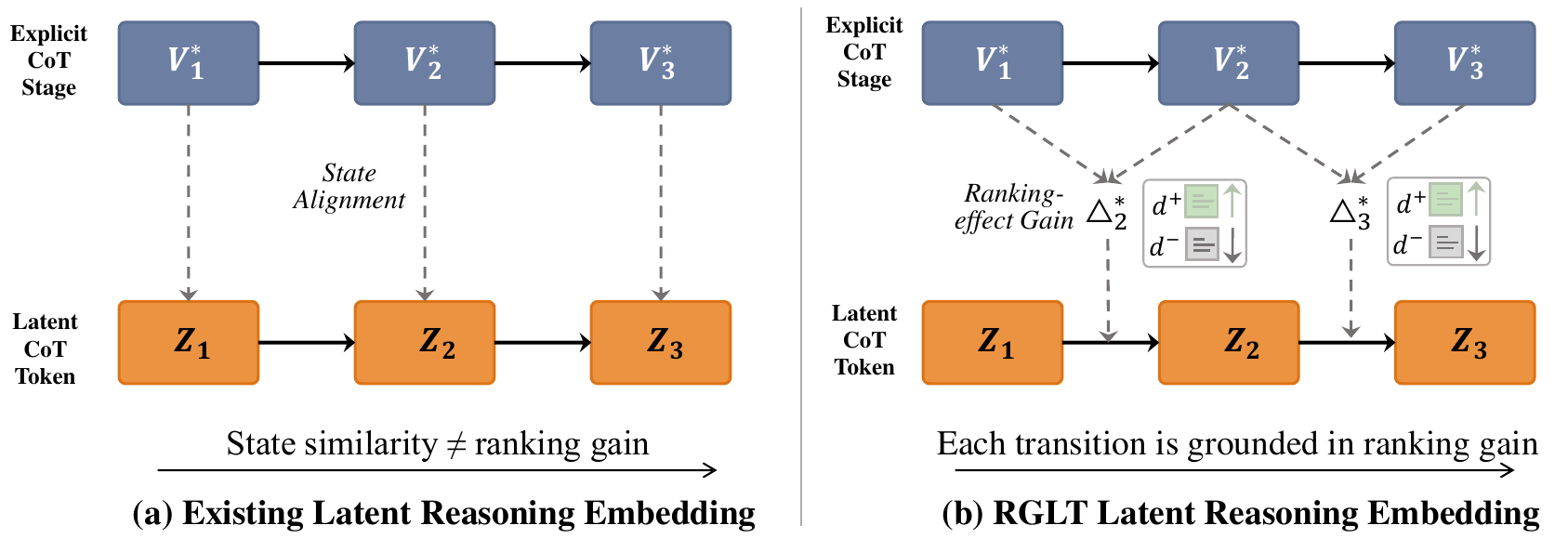}
\caption{Comparison of latent reasoning supervision. Existing methods supervise latent states through state alignment, which does not guarantee retrieval gains during reasoning evolution. RGLT instead transfers stage-wise ranking effects $\Delta^*_k$ from explicit CoT to latent transitions, ensuring that each reasoning step contributes to retrieval improvement.}
\label{Fig.1}
\end{figure}

However, constructing a latent reasoning trajectory does not necessarily translate into retrieval-relevant reasoning information. Existing approaches typically supervise latent reasoning through final retrieval objectives, distillation on final representations, or alignment between intermediate latent and explicit states (Figure~\ref{Fig.1}a). However, these objectives mainly optimize final retrieval outcomes or representation similarity, without directly modeling whether intermediate state transitions produce incremental retrieval gains. As a result, latent reasoning trajectories may learn shortcut reasoning patterns that maintain retrieval quality without contributing meaningful retrieval improvements across reasoning stages. The core challenge is therefore to shape latent reasoning trajectories whose state transitions progressively improve retrieval discrimination between relevant documents and hard negatives, rather than merely producing plausible intermediate
representations.

To address this challenge, we propose Retrieval Grounding Latent Reasoning (RGLT), a representation framework that performs non-autoregressive reasoning directly in latent space. RGLT appends a fixed sequence of silent tokens to construct an ordered latent reasoning trajectory, while a retrieval-intent anchor injects retrieval-oriented guidance into the trajectory based on the query and retrieval instruction. During training, explicit CoT is divided into multiple reasoning stages and provides cumulative semantic supervision for the corresponding latent states. Direct contrastive supervision further maintains the retrieval discriminability of each stage. More importantly, RGLT assigns stage-level credit according to how much each explicit CoT stage improves the retrieval preference for relevant documents over hard negatives. These retrieval gains are then propagated to the corresponding latent transitions, enabling the latent reasoning trajectory to learn retrieval-effective multi-step reasoning without enforcing direct representation alignment between the explicit and latent trajectories (Figure~\ref{Fig.1}b). The terminal latent state is used directly as the query embedding, requiring neither autoregressive CoT generation nor additional pooling or projection at inference.

Our contributions are summarized as follows:
\begin{itemize}
\item We identify an underexplored limitation of latent reasoning retrievers: intermediate transitions are not explicitly supervised for their incremental retrieval effects.

\item We introduce stage-level retrieval-effect credit assignment. Each latent transition must recover the ranking gain induced by its corresponding explicit CoT stage, ensuring that intermediate reasoning is grounded in actual retrieval improvements.

\item We propose RGLT, a non-autoregressive latent reasoner that surpasses autoregressive counterparts on three reasoning-intensive benchmarks, substantially reducing inference latency.
\end{itemize}

\section{Related Work}

\subsection{Explicit Reasoning for Dense Retrieval}

Recent work shows that LLM-generated reasoning can substantially improve retrieval for such settings. BRIGHT~\cite{bright} demonstrates the benefit of explicit reasoning on reasoning-intensive benchmarks. Data-centric methods, including ReasonIR~\cite{ReasonIR}, RaDeR~\cite{RaDeR}, and ReasonEmbed~\cite{reasonembed}, construct reasoning-oriented training instances, while Search-R3~\cite{Search-r3} and GRACE~\cite{grace} explicitly generate reasoning before forming the retrieval representation.

Meanwhile, LLM-based retrievers improve dense retrieval through instruction tuning, synthetic supervision, and joint generative--representational learning~\cite{wang-etal-2024-improving-text,ICLR2025_70cf2154,qwen3embedding}, while still encoding queries in a single forward pass. Query expansion methods instead externalize reasoning into text: HyDE~\cite{HyDE} generates a hypothetical document, Query2doc~\cite{query2doc} produces a pseudo-document, and DIVER~\cite{DIVER} iteratively refines the query with reasoning and retrieved evidence. Despite their effectiveness, these methods introduce autoregressive latency and make retrieval quality dependent on the generated
text.

\subsection{Latent Reasoning for Dense Retrieval}

Recent work has explored replacing textual reasoning with latent reasoning performed directly in hidden space. Coconut~\cite{coconut} iteratively feeds hidden states back as continuous thoughts, while CODI~\cite{codi} distills explicit CoT into continuous representations. In retrieval settings, GIRCSE~\cite{GIRCSE} generates soft embedding tokens, multi-query retrieval captures different relevance facets~\cite{AMER}, and AdaQR~\cite{AdaQR} approximates query reasoning through embedding-space transformations. However, these methods do not explicitly model whether intermediate latent transitions produce incremental retrieval gains.

Among existing retrieval-oriented latent reasoning methods, LaSER~\cite{laser}, which aligns explicit and latent trajectories, is most closely related to our work. While this constrains intermediate states, it ignores whether transitions between them yield actual retrieval gains. Instead, our method transfers stage-specific retrieval effects from explicit CoT to latent transitions, optimizing intermediate steps directly for retrieval improvements rather than mere state alignment or final-outcome optimization.

\section{Methodology}
\label{sec:method}

\begin{figure}[!t]
\centering
\includegraphics[width=\columnwidth]{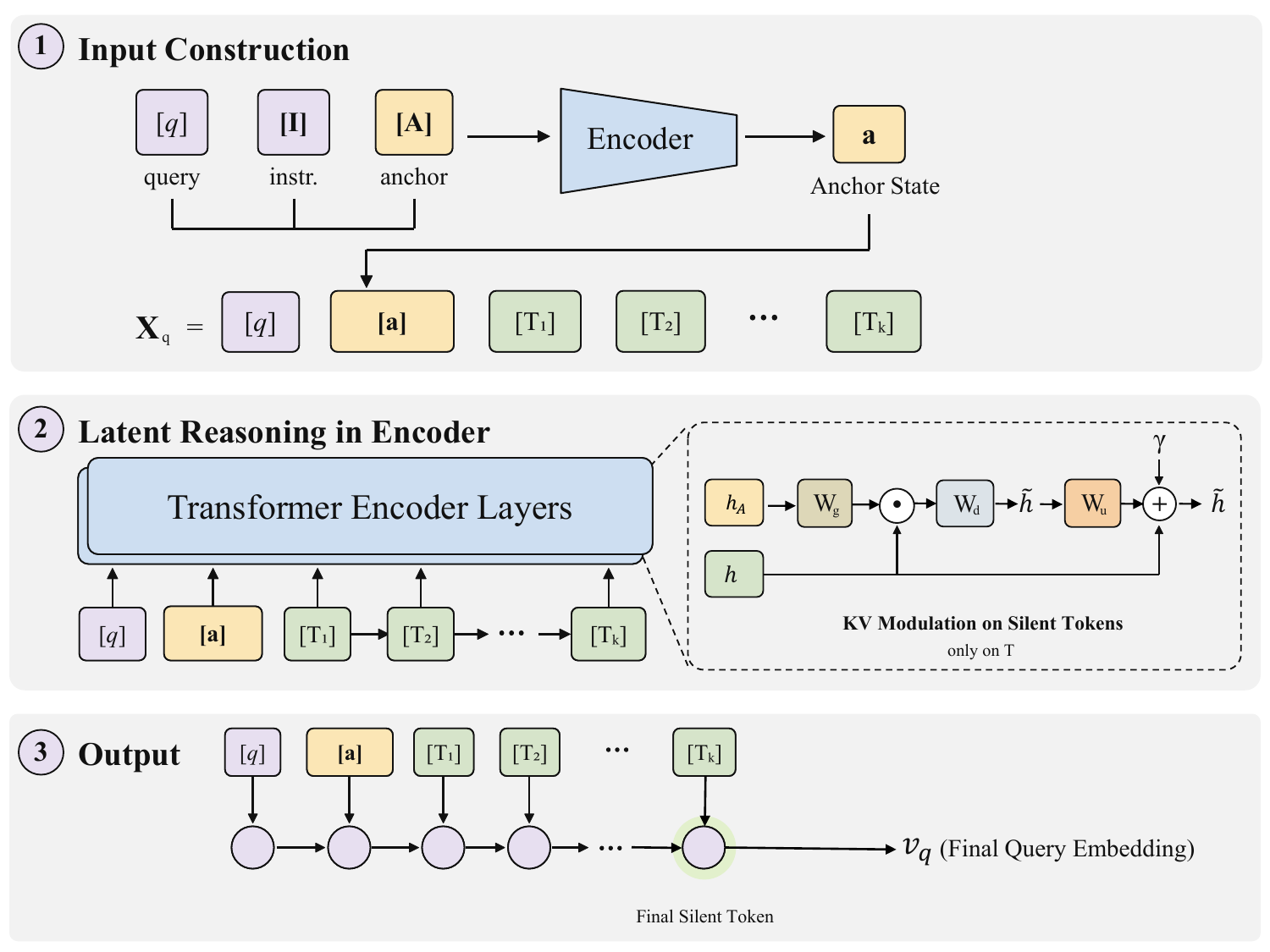}
\caption{Architecture of instruction-conditioned latent reasoning. The query and instruction are first compressed into a unified anchor state. During encoding, the anchor explicitly modulates the key and value representations of appended silent tokens via a gated residual to guide reasoning. Finally, the hidden state of the terminal silent token is directly extracted as the final query embedding.}
\label{fig:latent_reasoning}
\end{figure}

\begin{figure*}[t]
\centering
\includegraphics[width=\textwidth]{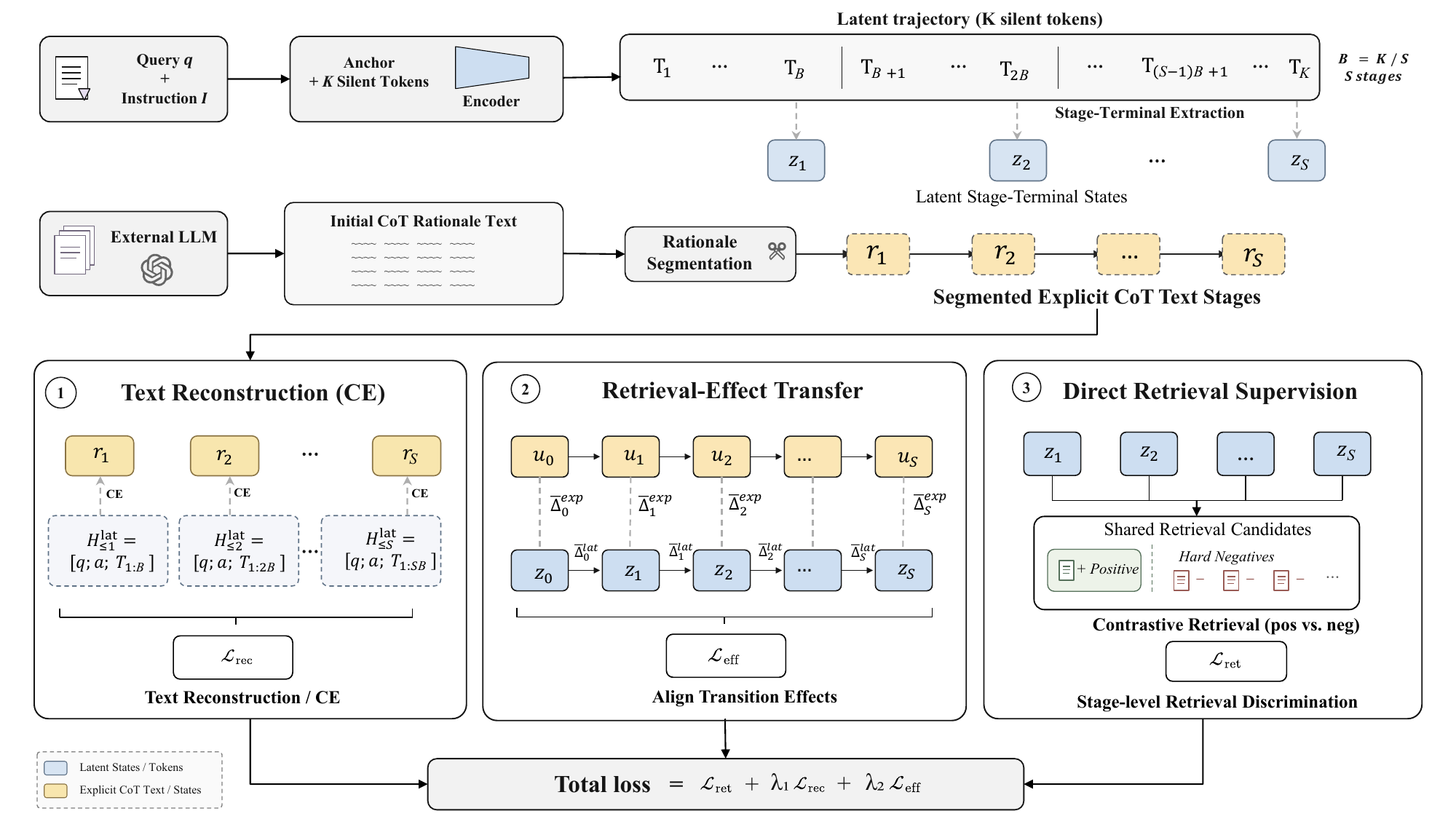}
\caption{The training pipeline of RGLT. It illustrates the generation of the latent reasoning trajectory and the subsequent joint optimization using three complementary losses: text reconstruction loss ($\mathcal{L}_{\mathrm{rec}}$), retrieval-effect transfer loss ($\mathcal{L}_{\mathrm{eff}}$), and direct retrieval supervision loss ($\mathcal{L}_{\mathrm{ret}}$).}
\label{pipeline}
\end{figure*}

In this section, we introduce the proposed RGLT in detail, whose overall architecture is illustrated in Figure~\ref{fig:latent_reasoning} and whose training pipeline is illustrated in Figure~\ref{pipeline}. RGLT appends a fixed number of latent reasoning tokens to the end of the input sequence, forming an implicit CoT trajectory in a non-autoregressive manner. During training, both the reasoning signals encoded in explicit CoT and the retrieval effects induced by its individual reasoning stages jointly guide the evolution of the latent reasoning tokens. Stage-level retrieval supervision further requires each reasoning stage to yield tangible gains in document ranking.

\subsection{Problem Formulation}
\label{sec:problem_formulation}

Given a user query $q$, a retrieval instruction $I$, and a candidate document corpus $\mathcal{D}=\{d_1,\ldots,d_N\}$, dense retrieval learns an encoder $f_\theta(\cdot)$ that maps queries and documents into a shared $m$-dimensional embedding space $\mathbb{R}^{m}$. Retrieval is then performed by comparing the query and document embeddings using cosine similarity: $s(q,d) = \cos(v_q, v_d)$, where $v_q = f_\theta([q;I])$ and $v_d = f_\theta(d)$ denote the query and document embeddings, and $[\cdot;\cdot]$ denotes concatenation under a fixed input template.

Reasoning-enhanced embedding models extend standard dense retrieval by allowing the model to construct a CoT (chain-of-thought) trajectory before forming the final query embedding:
\begin{equation}
[q;I] \longrightarrow r_1\longrightarrow\cdots\longrightarrow r_L \longrightarrow v_q^* .
\label{eq:cot_embedding}
\end{equation}
The intermediate reasoning units $r_1,\ldots,r_L$ may take the form of explicit tokens generated autoregressively or latent states evolving in continuous space. Such multi-stage reasoning can improve retrieval by progressively refining the query representation. However, autoregressive reasoning increases decoding latency and makes the resulting retrieval representation sensitive to the lexical structure of the generated reasoning trajectory, reducing the efficiency advantages of embedding-based retrieval.

In this work, we construct the latent reasoning trajectory directly within the encoder by appending $K$ fixed silent tokens, $\mathsf{T}_1,\ldots,\mathsf{T}_K$, to the instruction-conditioned query:
\begin{equation}
[q;I;\mathsf{T}_1;\ldots;\mathsf{T}_K]
\xrightarrow{f_\theta} v_q^* .
\label{eq:credit_formulation}
\end{equation}
The contextual states associated with these silent tokens form an implicit reasoning trajectory, where successive latent transitions progressively refine the retrieval representation. The entire latent reasoning process is completed within a single forward pass, avoiding long autoregressive decoding while maintaining efficient embedding inference.

\subsection{Instruction-Conditioned Latent Reasoning Trajectory}
\label{sec:latent_reasoning}
Although the non-autoregressive latent reasoning process is defined over a fixed latent trajectory, the silent positions still need to be conditioned on the retrieval instruction. This conditioning is necessary because the same query may correspond to different retrieval objectives under different instructions. To keep the latent reasoning trajectory aligned with the intended retrieval objective, we compress the query and instruction into a unified contextual anchor. Specifically, we introduce a special anchor token $\mathbf{A}=\texttt{<anchor>}$ and first derive its hidden representation from the instruction-conditioned query context:
\begin{equation}  
a = \operatorname{Encoder}([q; I; \mathbf{A}]).  \label{eq:anchor_conditioning}  
\end{equation}
Subsequently, when formally constructing the latent reasoning sequence, we remove the instruction $I$ from the sequence, directly inject the extracted vector $a$ as the input embedding of the anchor, and append $K$ fixed silent tokens $\mathsf{T}$:
\begin{equation}  X_q = [q; a; \mathsf{T}_1; \ldots; \mathsf{T}_K].  
\label{eq:latent_input}  \end{equation}
With this structure, the retrieval instruction influences the latent reasoning trajectory primarily through the anchor representation $a$, which serves as the main conditioning signal for the subsequent latent transitions. To make the anchor representation consistently affect the evolution of the silent tokens, we modulate their attention $K$ and $V$ representations at each layer. Let $h$ denote the original $K$ or $V$ vector of a silent token in the current layer. We then use the anchor hidden state $h_{\mathbf{A}}$ from the same layer to produce an updated representation $\widetilde{h}$ through a gated low-rank residual:
\begin{equation}
\widetilde{h} = h + \gamma W_u \left[ (W_d h) \odot \sigma(W_g h_{\mathbf{A}} + b) \right].
\label{eq:anchor_operator}
\end{equation}
Here, $W_g$ and $b$ parameterize the gating function that produces a channel-wise modulation signal, while $W_d$ and $W_u$ implement low-rank projections to reduce additional computation. The anchor representation therefore influences how information is propagated through the latent reasoning trajectory across layers. To preserve the reusability of document embeddings in dual-encoder retrieval, this modulation is applied only during query encoding and remains disabled for document representations.

The silent tokens accumulate contextual information sequentially, allowing the final silent token $\mathsf{T}_K$ to summarize the latent reasoning
trajectory without additional pooling or projection layers. We therefore use its hidden state as the final query embedding:
\begin{equation}
v_q = f_\theta(X_q, \mathsf{T}_K).
\label{eq:final_vq}
\end{equation}

\subsection{Process-Supervised Explicit-to-Implicit Reasoning Distillation}
\label{sec:stage_supervision}
Despite the expressive capacity of silent tokens for latent reasoning, training only with the final retrieval objective provides little signal for intermediate reasoning stages. As a result, latent reasoning may converge to shortcut reasoning patterns that maintain retrieval performance through shallow matching cues rather than retrieval-effective state transitions. To mitigate this issue, we introduce Process-Supervised Explicit-to-Implicit Reasoning Distillation, which uses explicit CoT to shape the latent reasoning trajectory and imposes stage-wise reconstruction constraints on the intermediate latent states.

For each training query $q$, we use an external LLM to generate an explicit multi-stage CoT rationale, from which $S$ ordered reasoning stages $(r_1, r_2, \dots, r_S)$ are extracted via regular expression matching without hard truncation. We further construct cumulative reasoning prefixes $r_{\leq k}$ by concatenating the first $k$ stages. These cumulative prefixes serve as intermediate reconstruction targets for the latent reasoning trajectory.

We divide the $K$ silent tokens into $S$ latent reasoning stages with an equal stage size $B = K/S$. At the end of the $k$-th latent stage, the corresponding latent states are required to reconstruct the cumulative reasoning prefix $r_{\leq k}$. We optimize this objective using a teacher-forced token-level cross-entropy loss:
\begin{equation}
\mathcal{L}_{\mathrm{rec}} =
\frac{1}{S}\sum_{k=1}^{S}
\operatorname{CE}_\theta\!\left(
r_{\leq k}
\mid
[q; \mathbf{A}(a); \mathsf{T}_{1:k \times B}]
\right).
\label{eq:reconstruction}
\end{equation}

To avoid information leakage, only the silent tokens up to stage $k \times B$ are exposed when computing the reconstruction loss for $r_{\leq k}$. This process-supervised reconstruction objective encourages the latent reasoning trajectory to capture progressively richer reasoning information across stages.

\subsection{Retrieval-Grounded Supervision for Latent Reasoning}
\label{sec:retrieval_credit}
Semantic reconstruction encourages latent stages to capture progressively richer reasoning information, but does not directly constrain whether intermediate states and transitions improve retrieval discrimination. We therefore introduce retrieval-grounded supervision from two complementary perspectives: direct stage supervision, which preserves retrieval discrimination throughout the latent reasoning trajectory, and retrieval-effect credit, which models the incremental retrieval gains induced by adjacent reasoning transitions.

\textbf{Direct stage supervision.}
A latent reasoning trajectory should maintain retrieval discrimination throughout the reasoning process rather than only at the final stage. Relying solely on final-state supervision risks leaving early latent steps unconstrained, which may cause intermediate representations to drift outside the meaningful retrieval space. To encourage retrieval-effective intermediate states, we apply retrieval supervision to both the final representation and the intermediate latent stages. Let $\mathcal{B}$ be the candidate pool, $P_q\subseteq\mathcal{B}$ its positive subset, and $\tau$ the contrastive temperature. For a query state $v$, we define the multi-positive contrastive loss and the overall retrieval objective as:
\begin{equation}
    R(v;\mathcal{B}) =
-\log
\frac{\sum_{d\in P_q}\exp(s(v,d)/\tau)}
{\sum_{d\in\mathcal{B}}\exp(s(v,d)/\tau)},
\end{equation}
\begin{equation}
\begin{aligned}
\mathcal{L}_{\mathrm{ret}}
=
R(v_S;\mathcal{B})
+\alpha R(v_S^*;\mathcal{B})
+\beta\sum_{k=1}^{S}w_kR(v_k;\mathcal{B})
\end{aligned}
\end{equation}
where $S$ is the number of reasoning stages, $v_S$ is the final embedding representation generated from the latent trajectory, and $v_S^*$ is the corresponding explicit representation (Hereafter, the superscript $*$ denotes states and associated quantities derived from the explicit Chain-of-Thought (CoT) trajectory). The non-negative stage weights $w_k$ sum to one. The first two terms optimize the final latent and explicit retrieval representations, while the last term encourages intermediate latent stages to preserve retrieval discrimination throughout the reasoning trajectory. 

\textbf{Retrieval-effect credit.}
Supervision on intermediate latent states alone does not capture whether a state transition produces meaningful retrieval gains. To explicitly model the retrieval effect of each reasoning transition, we measure how the retrieval preference over relevant documents and hard negatives changes between adjacent reasoning stages. Rather than relying on raw similarity differences, which can be trivially amplified without improving retrieval discrimination, we compute transition gains within a normalized local document distribution defined over a subset $\mathcal{S}\subseteq\mathcal{B}$ containing the positive document and several hard negatives:
\begin{equation}
    p_v(d) = \frac{\exp(s(v,d)/\tau_d)} {\sum_{d'\in\mathcal{S}}\exp(s(v,d')/\tau_d)},
\end{equation}
where $\tau_d$ is the distillation temperature. This distribution compresses unbounded similarity scores into the $(0, 1)$ interval and forces the probabilities over the local subset $\mathcal{S}$ to sum to 1. This competitive normalization increases retrieval discrimination by forcing gains on relevant documents to come at the expense of hard negatives within the same candidate set. Based on this relative confidence space, the latent and explicit credits of stage $k$ are strictly defined as:
\begin{equation}
   \begin{aligned} \Delta_k(d) &=\log p_{v_k}(d) -\operatorname{sg}\!\left[\log p_{v_{k-1}}(d)\right],\\ \Delta_k^*(d) &=\log p_{v_k^*}(d)-\log p_{v_{k-1}^*}(d), \end{aligned} 
\end{equation}
where the stop-gradient operator $\operatorname{sg}[\cdot]$ forcibly fixes the preceding latent state as a comparison baseline. This prevents the model from artificially increasing the transition gain by degrading the preceding state, forcing the incremental retrieval effect to be attributed to the current state $v_k$. Meanwhile, the explicit increment $\Delta_k^*(d)$ calculated from external CoT data officially serves as the ``effect target'' for learning here.

Let $\pi_k(d)=\operatorname{sg}[p_{v_{k-1}^*}(d)]$ be the distribution of the explicit preceding state. We zero-mean center both credits under $\pi_k$ to eliminate global shifts independent of the candidate documents:
\begin{equation}
    \begin{aligned} \overline{\Delta}_k(d) &=\Delta_k(d)-\mathbb{E}_{d'\sim\pi_k}[\Delta_k(d')],\\ \overline{\Delta}_k^*(d) &=\Delta_k^*(d)-\mathbb{E}_{d'\sim\pi_k}[\Delta_k^*(d')]. \end{aligned}
\end{equation}
The final effect loss is:
\begin{equation}
    \mathcal{L}_{\mathrm{eff}} = \sum_{k=1}^{S} \mathbb{E}_{d\sim\pi_k} \left[ \left( \overline{\Delta}_k(d) -\operatorname{sg}[\overline{\Delta}_k^*(d)] \right)^2 \right].
\end{equation}
$\mathcal{L}_{\mathrm{eff}}$ transfers the stage-specific retrieval effects induced by explicit CoT reasoning to the corresponding latent transitions. Rather than constraining latent states to match explicit representations, the objective focuses on whether each latent transition produces comparable retrieval gains over relevant documents and hard negatives. This encourages the latent reasoning trajectory to learn retrieval-effective state transitions that progressively improve retrieval discrimination across reasoning stages.

\subsection{Objective Function}
\label{sec:objective_function}

The complete optimization objective is defined as:
\begin{equation}
\mathcal{L}
=\mathcal{L}_{\mathrm{ret}}
+\lambda_1\mathcal{L}_{\mathrm{rec}}
+\lambda_2\mathcal{L}_{\mathrm{eff}},
\label{eq:overall_objective}
\end{equation}
where \(\lambda_1\) and \(\lambda_2\) balance semantic reconstruction and retrieval-effect credit. These auxiliary weights are ramped up from zero early in training, allowing the encoder to establish a stable document space before enforcing latent trajectory constraints.

\begin{table*}[t]
\centering
\caption{Overall performance on reasoning-intensive retrieval benchmarks.
All values are percentages.}
\label{tab:main_results}
\scriptsize
\setlength{\tabcolsep}{2.6pt}

\begin{tabular*}{\textwidth}{
@{\extracolsep{\fill}}
lccccccccccc
@{}
}
\toprule
\multirow{3}{*}{Model}
& \multirow{3}{*}{Size}
& \multicolumn{2}{c}{BRIGHT}
& \multicolumn{5}{c}{FollowIR}
& \multicolumn{3}{c}{BrowseComp-Plus} \\
\cmidrule(lr){3-4}
\cmidrule(lr){5-9}
\cmidrule(lr){10-12}

& & nDCG@10 & R@10
& Robust04 & News21 & Core17
& Avg. & Avg.
& R@5 & R@100 & R@1000 \\

& & & 
& MAP@5 & nDCG@5 & MAP@5
& Score & p-MRR
& & & \\
\midrule

\multicolumn{12}{l}{\emph{Standard dense retrievers}} \\

BGE-M3
& 0.6B
& 11.40 & 14.39
& 1.50 & 21.40 & 7.50
& 10.10 & -3.90
& 4.10 & 21.70 & 47.20 \\

E5-Mistral-7B-Instruct
& 7B
& 16.50 & 20.12
& 2.70 & 28.80 & 13.70
& 15.10 & -1.50
& 9.30 & 37.10 & 70.20 \\

Qwen3-Embedding-8B
& 8B
& 14.00 & 17.31
& 3.10 & 25.50 & 10.10
& 12.90 & 7.20
& 7.70 & 31.60 & 61.30 \\

\midrule
\multicolumn{12}{l}{\emph{Basic contrastive training}} \\

Fair Baseline (Qwen3-0.6B)
& 0.6B
& 18.30 & 22.14
& 2.10 & 13.40 & 6.70
& 7.40 & -0.20
& 3.50 & 21.30 & 46.90 \\

Fair Baseline (Qwen3-8B)
& 8B
& 25.70 & 30.45
& 2.80 & 18.90 & 11.20
& 11.00 & 1.70
& 11.30 & 37.40 & 63.20 \\

Fair Baseline (LLaMA3.1-8B)
& 8B
& 22.50 & 26.86
& 2.50 & 18.90 & 8.10
& 9.80 & 0.10
& 6.10 & 25.40 & 50.80 \\

\midrule
\multicolumn{12}{l}{\emph{Explicit reasoning}} \\

Rewrite-then-Retrieve (Qwen3-8B)
& 8B
& 28.10 & 33.15
& -- & -- & --
& -- & --
& -- & -- & -- \\

Search-R3 (Qwen2.5-1.5B)
& 1.5B
& 7.70 & 10.24
& 3.20 & 26.20 & 10.10
& 13.20 & 3.20
& 0.00 & 0.30 & 1.10 \\

InBedder (LLaMA2-7B)
& 7B
& 5.56 & 7.50
& 2.60 & 8.50 & 1.60
& 4.20 & -0.10
& 1.40 & 8.10 & 27.40 \\

\midrule
\multicolumn{12}{l}{\emph{Latent reasoning}} \\

GIRCSE (Qwen3-8B)
& 8B
& 26.00 & 30.79
& 3.00 & 22.60 & 8.50
& 11.40 & 2.00
& 13.00 & 40.80 & 68.10 \\

LaSER (Qwen3-8B)$^\star$
& 8B
& 29.90 & 34.27
& 4.10 & 21.80 & \textbf{11.40}
& 12.50 & 2.40
& 11.70 & 38.40 & 66.90 \\

\textsc{RGLT} (Qwen3-8B)
& 8B
& \textbf{34.20} & \textbf{39.97}
& \textbf{12.90} & \textbf{23.60} & 4.60
& \textbf{13.70} & \textbf{2.62}
& \textbf{13.42} & \textbf{41.8} & \textbf{70.83} \\

\bottomrule
\end{tabular*}

\vspace{2pt}
\parbox{\textwidth}{\scriptsize
Rewrite-then-Retrieve uses an external LLM to generate a
reasoning-enhanced query before retrieval.}
\end{table*}

\section{Experiments}
\label{sec:experiments}

We organize the experiments around four research questions. \emph{RQ1: Effectiveness.} Does \textsc{RGLT} improve reasoning-intensive retrieval over standard, explicit-reasoning, and latent-reasoning retrievers? \emph{RQ2: Components.} Which parts of \textsc{RGLT} account for the improvement? \emph{RQ3: Representation evolution.} Do the latent stages yield progressively stronger retrieval representations? \emph{RQ4: Efficiency and sensitivity.} What is the query-side cost, and how sensitive is the method to the silent-token budget?

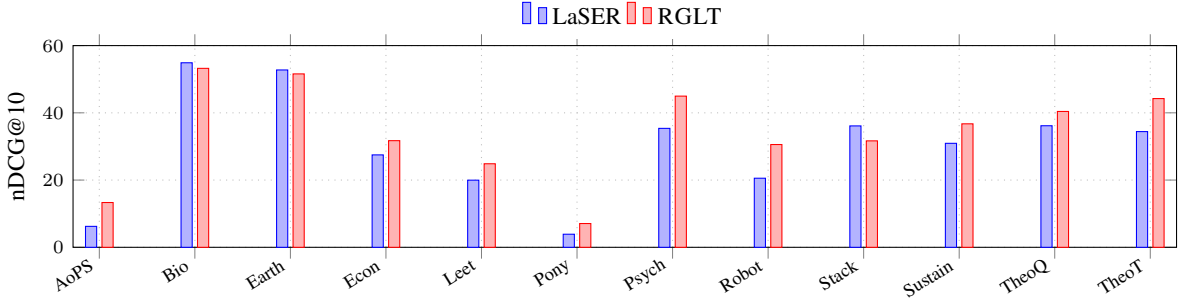
\begin{figure*}[t]
\centering
\begin{tikzpicture}
\begin{axis}[
    ybar,
    bar width=4.2pt,
    width=0.98\textwidth,
    height=4.25cm,
    ymin=0,
    ymax=60,
    ylabel={nDCG@10},
    symbolic x coords={AoPS,Bio,Earth,Econ,Leet,Pony,Psych,Robot,Stack,Sustain,TheoQ,TheoT},
    xtick=data,
    x tick label style={rotate=32,anchor=east,font=\scriptsize},
    y tick label style={font=\scriptsize},
    ylabel style={font=\small},
    legend style={at={(0.5,1.03)},anchor=south,legend columns=2,font=\small,draw=none},
    enlarge x limits=0.025,
    grid=major,
    major grid style={dotted},
]
\addplot coordinates {
(AoPS,6.23) (Bio,54.89) (Earth,52.75) (Econ,27.49)
(Leet,19.97) (Pony,3.89) (Psych,35.38) (Robot,20.55)
(Stack,36.11) (Sustain,30.93) (TheoQ,36.16) (TheoT,34.42)};
\addplot coordinates {
(AoPS,13.32) (Bio,53.23) (Earth,51.58) (Econ,31.72)
(Leet,24.85) (Pony,7.07) (Psych,44.97) (Robot,30.57)
(Stack,31.66) (Sustain,36.73) (TheoQ,40.41) (TheoT,44.24)};
\legend{LaSER,\textsc{RGLT}}
\end{axis}
\end{tikzpicture}
\caption{Domain-level BRIGHT nDCG@10. \textsc{RGLT} outperforms LaSER on 9 of 12 domains.}
\label{fig:domain_results}
\end{figure*}

\subsection{Experimental Setup}
\label{sec:experimental_setup}
\paragraph{Datasets and metrics.}
We train on the 81,659 instances from ReasonEmbed~\cite{reasonembed}, where each query is paired with relevant documents, hard negatives, and an explicit reasoning path. We evaluate on BRIGHT~\cite{bright}, FollowIR~\cite{followir}, and BrowseComp-Plus~\cite{browsecomp}. BRIGHT contains 1,384 queries and 1,145,164 documents from 12 technical and scientific domains; we report nDCG@10 and Recall@10. For FollowIR, we report the averaged task score and p-MRR. For BrowseComp-Plus, we report Recall@5, Recall@100, and Recall@1000.

\paragraph{Baselines.}
We compare with four groups of methods. Standard retrievers include BGE-M3~\cite{bge-m3}, E5-Mistral-7B-Instruct~\cite{E5-mistral-7B-instruct}, and Qwen3-Embedding-8B~\cite{qwen3embedding}. The \emph{Fair Baseline} fine-tunes the same Qwen3-8B backbone and training data using only contrastive learning. Explicit reasoning methods include Rewrite-then-Retrieve, Search-R3~\cite{Search-r3}, and InBedder~\cite{InBedder}. Implicit reasoning methods include GIRCSE~\cite{GIRCSE} and LaSER~\cite{laser}. LaSER is closely related to our work because it connects explicit CoT
reasoning with latent retrieval representations through intermediate
trajectory supervision.

\paragraph{Implementation details.}
We use Qwen3-8B. The $K$ silent tokens form four cumulative $K/4$ stages, and the hidden state of $\mathsf{T}_K$ serves as the query embedding (document encoding omits silent tokens). We train for three epochs (LoRA rank 32) via Hugging Face Accelerate (torch.distributed) on an internal Linux cluster of 32 GPUs (4 $\times$ 8 A100, 64GB RAM/worker; Docker: Python 3.9, PyTorch 2.3.1, CUDA 12.4.0, FlashAttention 2.5.8, NCCL 2.20.5, OpenMPI 4, GCC 11). A per-device query batch of 4 (global 128) pairs 1 positive with 3 hard negatives per query, forming a 512-document pool via cross-device gathering. Online-generated explicit CoT states from the shared backbone act as stop-gradient targets. BRIGHT evaluation spans all 12 domains (1,384 queries) with fresh document embeddings and retrieval instructions enabled. With auxiliary weights optimally set to $\lambda_1=\lambda_2=0.10$, all reported results are averaged over three independent runs.

\subsection{Main Results}
\label{sec:main_results}
\emph{RQ1: Effectiveness.}
Table~\ref{tab:main_results} shows that \textsc{RGLT} achieves the best completed BRIGHT result, with 34.20 nDCG@10 and 39.97 Recall@10. It improves over LaSER by 14.4\% and 16.6\% respectively, and exceeds the Fair Baseline by 8.50 nDCG points. \textsc{RGLT} also outperforms GIRCSE and Rewrite-then-Retrieve by 8.20 and 6.10 nDCG points. These results show that retrieval-effect supervision is more effective than contrastive-only latent refinement and avoids explicit CoT generation at inference.

Figure~\ref{fig:domain_results} shows that the gain is broad rather than domain-specific. \textsc{RGLT} improves 9 of 12 domains, with clear gains on AoPS, psychology, robotics, and both TheoremQA subsets. It also improves LeetCode, economics, and sustainable living. LaSER remains better on biology, earth science, and StackOverflow, but the gaps on the first two are small. Moreover, \textsc{RGLT} obtains higher Recall@10 on 11 of 12 domains. The macro improvement therefore comes from more balanced retrieval across domains, not from a single outlier.

\begin{table}[t]
\centering
\caption{Component ablations and architectural variants on BRIGHT. The lower block contains independently trained variants and is not a one-factor ablation.}
\label{tab:ablation}
\scriptsize
\setlength{\tabcolsep}{4.0pt}
\begin{tabular}{lcc}
\toprule
Variant & nDCG@10 & R@10 \\
\midrule
\textsc{RGLT} & \textbf{34.20} & \textbf{39.97} \\
\quad w/o stage retrieval-effect matching & 31.26 & 36.15 \\
\quad w/o intent retrieval-effect matching & 31.69 & 36.85 \\
\quad w/o ST16-vs-base advantage & 26.39 & 32.56 \\
\quad replace effect matching with state alignment & 30.64 & 35.69 \\
\midrule
Gated retrieval-credit fusion variant & 32.14 & 37.94 \\
Naive terminal-ST readout variant & 30.87 & 37.01 \\
Reasoning-query distillation variant & 32.01 & 37.85 \\
\bottomrule
\end{tabular}
\end{table}


\subsection{Ablation and Architecture Analysis}
\label{sec:ablation}

\emph{RQ2: Components.}
Table~\ref{tab:ablation} first shows that a terminal silent state alone is insufficient. The naive terminal-ST variant reaches 30.87 nDCG@10, whereas \textsc{RGLT} reaches 34.20 by grounding the same stage terminals in explicit retrieval effects. The proposed stage-terminal design also outperforms the earlier gated-fusion variant, increasing nDCG@10 from 32.14 to 34.20 and Recall@10 from 37.94 to 39.97. Thus, the latent trajectory can directly serve as the retrieval representation without an external fusion module. All ablated variants underperform the full model. In particular, replacing retrieval-effect matching with state alignment causes a clear drop, confirming that modeling stage-wise ranking changes is more effective than aligning intermediate states alone.

\begin{table}[t]
\centering
\caption{BRIGHT performance of cumulative latent stages. ST4--ST16 are the terminal states of four-token stages.}
\label{tab:stage_results}
\scriptsize
\setlength{\tabcolsep}{5.2pt}
\begin{tabular}{lcc}
\toprule
Query representation & nDCG@10 & R@10 \\
\midrule
Base query state & 26.39 & 30.24 \\
Stage 1 terminal (ST4) & 32.25 & 34.96 \\
Stage 2 terminal (ST8) & 32.48 & 35.25 \\
Stage 3 terminal (ST12) & 33.56 & 37.14 \\
Stage 4 terminal (ST16) & \textbf{34.20} & \textbf{39.97} \\
\bottomrule
\end{tabular}
\end{table}

\subsection{Retrieval-Grounded Representation Evolution}
\label{sec:evolution_analysis}
\emph{RQ3: Representation evolution.}
Table~\ref{tab:stage_results} reports the retrieval performance of the base representation and the four stage-terminal states. nDCG@10 increases from 26.39 at the base state to 32.25, 32.48, 33.56, and 34.20 at ST4, ST8, ST12, and ST16, respectively. Recall@10 follows the same trend, increasing from 30.24 to 39.97.
These results show that retrieval utility accumulates progressively along the latent trajectory rather than emerging only at the final stage. The best performance at ST16 also supports its direct use as the final query embedding.

\subsection{Efficiency and Sensitivity Analysis}
\label{sec:efficiency}

\begin{table}[t]
\centering
\caption{Query-side inference efficiency on BRIGHT. Latency is
measured over 80 queries with batch size 8 on one NVIDIA A100
80GB GPU.}
\label{tab:efficiency}
\scriptsize
\setlength{\tabcolsep}{3.3pt}
\begin{tabular}{lcccc}
\toprule
Method & Latency & Relative & Text gen. & Index cost \\
       & (ms/query) &  &  &  \\
\midrule
Base retriever (Qwen3-8B)
    & $21.5$ & $1.00\times$ & No & None \\
Rewrite-then-Retrieve
    & $4000$ & $186\times$ & Yes & None \\
LaSER (Qwen3-8B)
    & $32.5$ & $1.51\times$ & No & None \\
\textsc{RGLT} (Qwen3-8B)
    & $\textbf{24.0}$ & $\textbf{1.12}\times$ & No & None \\
\bottomrule
\end{tabular}
\end{table}

\emph{RQ4: Efficiency and sensitivity.} Table~\ref{tab:efficiency} details the query-side inference efficiency across different methods. As shown, explicitly generating textual CoT (Rewrite-then-Retrieve) introduces a prohibitive $186\times$ latency overhead. In contrast, \textsc{RGLT} operates without generating textual CoT, leaves the document encoding unchanged, and fully preserves single-vector nearest-neighbor search. While LaSER also avoids text generation, its recurrent construction of latent tokens incurs a $1.51\times$ latency penalty. \textsc{RGLT} overcomes this by exposing all $K$ fixed silent positions simultaneously in a non-autoregressive forward pass, directly extracting the ST$K$ representation. Because the extra computation is strictly limited to a fixed query-side sequence extension, \textsc{RGLT} achieves a latency of 24.0 ms/query under identical measurement setups, representing a minimal 12\% overhead ($1.12\times$) over the base retriever.

\begin{figure}[h]
  \centering
  \includegraphics[width=0.62\linewidth]{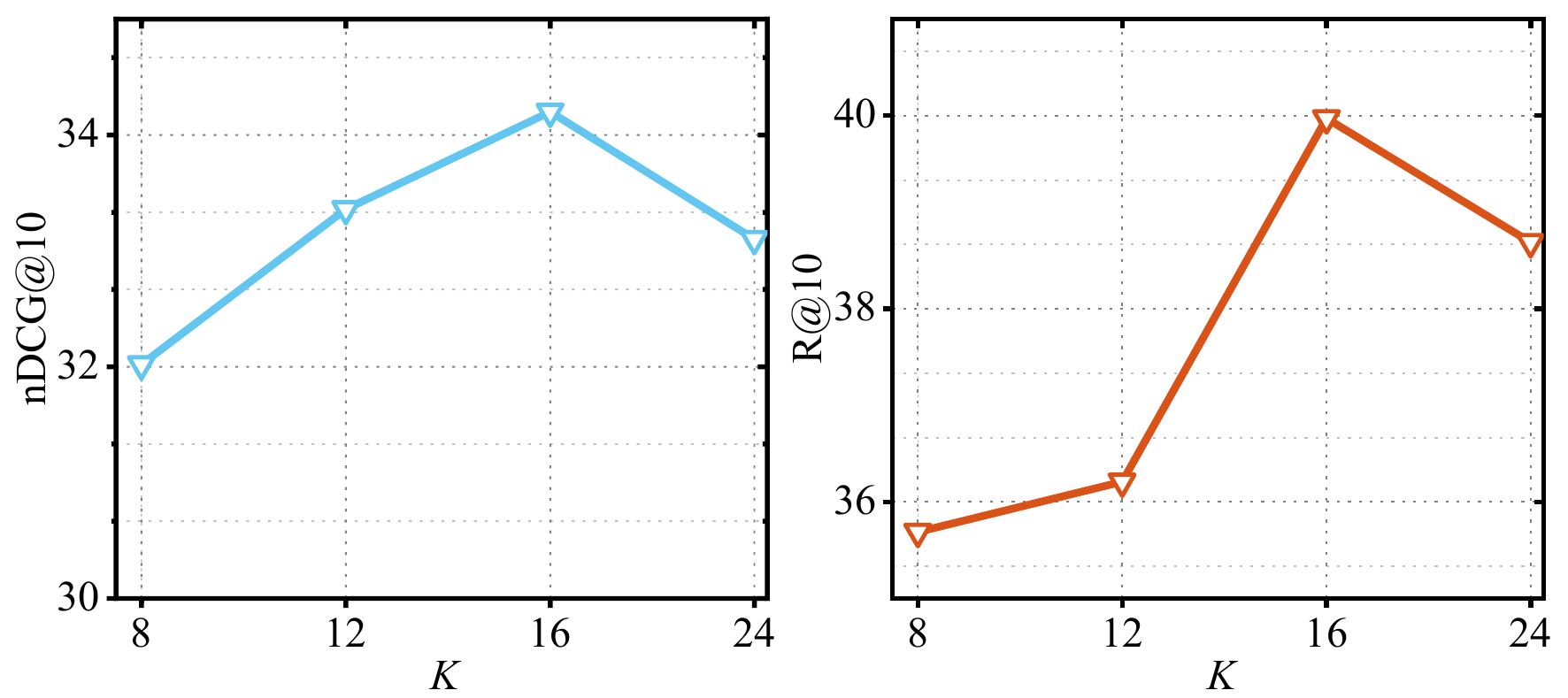}
  \caption{Impact of the latent budget ($K$) on retrieval performance.}
\label{fig:parameter}
\end{figure}

Figure~\ref{fig:parameter} evaluates retrieval performance across different latent budgets ($K$). Performance peaks at the default setting of $K=16$ (four tokens per stage) for both nDCG@10 and Recall@10. However, metrics drop at $K=24$, indicating that performance does not scale indefinitely with the latent budget; an excessively large $K$ may complicate optimization, making $K=16$ the optimal choice.

\section{Conclusion}

In this paper, we presented \textsc{RGLT}, a latent reasoning framework for reasoning-intensive dense retrieval. Unlike existing methods that supervise only the final representation or align intermediate states,
\textsc{RGLT} explicitly grounds each latent transition in the retrieval effect of its corresponding CoT stage. This enables the model to learn an ordered evolution of retrieval representations, while the terminal latent state can be used directly as a single-vector query embedding without autoregressive CoT generation, additional pooling, or document-side modification. Experiments on reasoning-intensive retrieval benchmarks show that \textsc{RGLT} consistently outperforms standard, explicit-reasoning, and latent-reasoning retrievers. Analyses confirm that retrieval
quality improves progressively across latent stages and that retrieval-effect matching is more effective than conventional state
alignment. These results show that latent reasoning becomes more effective for retrieval when its intermediate computation is grounded in document-ranking improvements.

\clearpage
\bibliography{aaai2027}

@inproceedings{wang-etal-2024-improving-text,
    title = "Improving Text Embeddings with Large Language Models",
    author = "Wang, Liang  and
      Yang, Nan  and
      Huang, Xiaolong  and
      Yang, Linjun  and
      Majumder, Rangan  and
      Wei, Furu",
    editor = "Ku, Lun-Wei  and
      Martins, Andre  and
      Srikumar, Vivek",
    booktitle = "Proceedings of the 62nd Annual Meeting of the Association for Computational Linguistics (Volume 1: Long Papers)",
    month = aug,
    year = "2024",
    address = "Bangkok, Thailand",
    publisher = "Association for Computational Linguistics",
    url = "https://aclanthology.org/2024.acl-long.642/",
    doi = "10.18653/v1/2024.acl-long.642",
    pages = "11897--11916"
}

@inproceedings{ICLR2025_70cf2154,
 author = {Muennighoff, Niklas and SU, Hongjin and Wang, Liang and Yang, Nan and Wei, Furu and Yu, Tao and Singh, Amanpreet and Kiela, Douwe},
 booktitle = {International Conference on Learning Representations},
 editor = {Y. Yue and A. Garg and N. Peng and F. Sha and R. Yu},
 pages = {45544--45613},
 title = {Generative Representational Instruction Tuning},
 url = {https://proceedings.iclr.cc/paper_files/paper/2025/file/70cf215430492f7d34830a24e744b3f1-Paper-Conference.pdf},
 volume = {2025},
 year = {2025}
}

@misc{qwen3embedding,
      title={Qwen3 Embedding: Advancing Text Embedding and Reranking Through Foundation Models}, 
      author={Yanzhao Zhang and Mingxin Li and Dingkun Long and Xin Zhang and Huan Lin and Baosong Yang and Pengjun Xie and An Yang and Dayiheng Liu and Junyang Lin and Fei Huang and Jingren Zhou},
      year={2025},
      eprint={2506.05176},
      archivePrefix={arXiv},
      primaryClass={cs.CL},
      url={https://arxiv.org/abs/2506.05176}, 
}

@inproceedings{HyDE,
    title = "Precise Zero-Shot Dense Retrieval without Relevance Labels",
    author = "Gao, Luyu  and
      Ma, Xueguang  and
      Lin, Jimmy  and
      Callan, Jamie",
    editor = "Rogers, Anna  and
      Boyd-Graber, Jordan  and
      Okazaki, Naoaki",
    booktitle = "Proceedings of the 61st Annual Meeting of the Association for Computational Linguistics (Volume 1: Long Papers)",
    month = jul,
    year = "2023",
    address = "Toronto, Canada",
    publisher = "Association for Computational Linguistics",
    url = "https://aclanthology.org/2023.acl-long.99/",
    doi = "10.18653/v1/2023.acl-long.99",
    pages = "1762--1777"
}

@inproceedings{query2doc,
    title = "Query2doc: Query Expansion with Large Language Models",
    author = "Wang, Liang  and
      Yang, Nan  and
      Wei, Furu",
    editor = "Bouamor, Houda  and
      Pino, Juan  and
      Bali, Kalika",
    booktitle = "Proceedings of the 2023 Conference on Empirical Methods in Natural Language Processing",
    month = dec,
    year = "2023",
    address = "Singapore",
    publisher = "Association for Computational Linguistics",
    url = "https://aclanthology.org/2023.emnlp-main.585/",
    doi = "10.18653/v1/2023.emnlp-main.585",
    pages = "9414--9423"
}

@misc{DIVER,
      title={{DIVER}: A Multi-Stage Approach for Reasoning-intensive Information Retrieval}, 
      author={Duolin Sun and Meixiu Long and Dan Yang and Junjie Wang and Yecheng Luo and Yue Shen and Jian Wang and Hualei Zhou and Chunxiao Guo and Peng Wei and Jiahai Wang and Jinjie Gu},
      year={2026},
      eprint={2508.07995},
      archivePrefix={arXiv},
      primaryClass={cs.IR},
      url={https://arxiv.org/abs/2508.07995}, 
}

@inproceedings{bright,
  title={{BRIGHT}: A realistic and challenging benchmark for reasoning-intensive retrieval},
  author={Su, Hongjin and Yen, Howard and Xia, Mengzhou and Shi, Weijia and Muennighoff, Niklas and Wang, Han-yu and Haisu, Liu and Shi, Quan and Siegel, Zachary and Tang, Michael and others},
  booktitle={International Conference on Learning Representations},
  volume={2025},
  pages={48941--48991},
  year={2025}
}

@inproceedings{ReasonIR,
  title={{ReasonIR}: Training Retrievers for Reasoning Tasks},
  author={Shao, Rulin and Qiao, Rui and Kishore, Varsha and Muennighoff, Niklas and Lin, Xi Victoria and Rus, Daniela and Low, Bryan Kian Hsiang and Min, Sewon and Yih, Wen-tau and Koh, Pang Wei and others},
  booktitle={Second Conference on Language Modeling},
  year={2025}
}

@inproceedings{followir,
    title = "{F}ollow{IR}: Evaluating and Teaching Information Retrieval Models to Follow Instructions",
    author = "Weller, Orion  and
      Chang, Benjamin  and
      MacAvaney, Sean  and
      Lo, Kyle  and
      Cohan, Arman  and
      Van Durme, Benjamin  and
      Lawrie, Dawn  and
      Soldaini, Luca",
    editor = "Chiruzzo, Luis  and
      Ritter, Alan  and
      Wang, Lu",
    booktitle = "Proceedings of the 2025 Conference of the Nations of the Americas Chapter of the Association for Computational Linguistics: Human Language Technologies (Volume 1: Long Papers)",
    month = apr,
    year = "2025",
    address = "Albuquerque, New Mexico",
    publisher = "Association for Computational Linguistics",
    url = "https://aclanthology.org/2025.naacl-long.597/",
    doi = "10.18653/v1/2025.naacl-long.597",
    pages = "11926--11942",
    ISBN = "979-8-89176-189-6"
}

@inproceedings{InBedder,
  title={Answer is all you need: Instruction-following text embedding via answering the question},
  author={Peng, Letian and Zhang, Yuwei and Wang, Zilong and Srinivasa, Jayanth and Liu, Gaowen and Wang, Zihan and Shang, Jingbo},
  booktitle={Proceedings of the 62nd Annual Meeting of the Association for Computational Linguistics (Volume 1: Long Papers)},
  pages={459--477},
  year={2024}
}

@inproceedings{browsecomp,
  title={{BrowseComp-Plus}: A more fair and transparent evaluation benchmark of deep-research agent},
  author={Chen, Zijian and Ma, Xueguang and Zhuang, Shengyao and Nie, Ping and Zou, Kai and Sharifymoghaddam, Sahel and Liu, Andrew and Green, Joshua and Patel, Kshama and Meng, Ruoxi and others},
  booktitle={First Workshop on Multi-Turn Interactions in Large Language Models},
  year={2025}
}

@article{E5-mistral-7B-instruct,
  title={Text embeddings by weakly-supervised contrastive pre-training},
  author={Wang, Liang and Yang, Nan and Huang, Xiaolong and Jiao, Binxing and Yang, Linjun and Jiang, Daxin and Majumder, Rangan and Wei, Furu},
  journal={arXiv preprint arXiv:2212.03533},
  year={2022}
}

@article{bge-m3,
  title={{M3-Embedding}: Multi-linguality, multi-functionality, multi-granularity text embeddings through self-knowledge distillation},
  author={Chen, Jianlv and Xiao, Shitao and Zhang, Peitian and Luo, Kun and Lian, Defu and Liu, Zheng},
  journal={arXiv preprint arXiv:2402.03216},
  year={2024}
}

@inproceedings{reasonembed,
    title = "{R}eason{E}mbed: Enhanced Text Embeddings for Reasoning-Intensive Document Retrieval",
    author = "Chen, Jianlyu  and
      Lan, Junwei  and
      Li, Chaofan  and
      Lian, Defu  and
      Liu, Zheng",
    editor = "Liakata, Maria  and
      Moreira, Viviane P.  and
      Zhang, Jiajun  and
      Jurgens, David",
    booktitle = "Proceedings of the 64th Annual Meeting of the {A}ssociation for {C}omputational {L}inguistics (Volume 1: Long Papers)",
    month = jul,
    year = "2026",
    address = "San Diego, California, United States",
    publisher = "Association for Computational Linguistics",
    url = "https://aclanthology.org/2026.acl-long.54/",
    doi = "10.18653/v1/2026.acl-long.54",
    pages = "1203--1221",
    ISBN = "979-8-89176-390-6"
}

@inproceedings{RaDeR,
    title = "{R}a{D}e{R}: Reasoning-aware Dense Retrieval Models",
    author = "Das, Debrup  and
      O{'}Nuallain, Sam  and
      Rahimi, Razieh",
    editor = "Christodoulopoulos, Christos  and
      Chakraborty, Tanmoy  and
      Rose, Carolyn  and
      Peng, Violet",
    booktitle = "Proceedings of the 2025 Conference on Empirical Methods in Natural Language Processing",
    month = nov,
    year = "2025",
    address = "Suzhou, China",
    publisher = "Association for Computational Linguistics",
    url = "https://aclanthology.org/2025.emnlp-main.1011/",
    doi = "10.18653/v1/2025.emnlp-main.1011",
    pages = "19970--19997",
    ISBN = "979-8-89176-332-6"
}

@article{Search-r3,
  title={{Search-R3}: Unifying reasoning and embedding generation in large language models},
  author={Gui, Yuntao and Cheng, James},
  journal={arXiv preprint arXiv:2510.07048},
  year={2025}
}

@article{grace,
  title={{GRACE}: Generative Representation Learning via Contrastive Policy Optimization},
  author={Sun, Jiashuo and Liu, Shixuan and Su, Zhaochen and Zhong, Xianrui and Jiang, Pengcheng and Jin, Bowen and Li, Peiran and Shi, Weijia and Han, Jiawei},
  journal={arXiv preprint arXiv:2510.04506},
  year={2025}
}

@inproceedings{coconut,
  title={Training Large Language Models to Reason in a Continuous Latent Space},
  author={Hao, Shibo and Sukhbaatar, Sainbayar and Su, DiJia and Li, Xian and Hu, Zhiting and Weston, Jason E and Tian, Yuandong},
  booktitle={Second Conference on Language Modeling},
  year={2025}
}

@inproceedings{codi,
  title={{CODI}: Compressing chain-of-thought into continuous space via self-distillation},
  author={Shen, Zhenyi and Yan, Hanqi and Zhang, Linhai and Hu, Zhanghao and Du, Yali and He, Yulan},
  booktitle={Proceedings of the 2025 Conference on Empirical Methods in Natural Language Processing},
  pages={677--693},
  year={2025}
}

@article{GIRCSE,
  title={Let {LLMs} speak embedding languages: Generative text embeddings via iterative contrastive refinement},
  author={Tsai, Yu-Che and Chen, Kuan-Yu and Li, Yuan-Chi and Chen, Yuan-Hao and Tsai, Ching-Yu and Lin, Shou-De},
  journal={arXiv preprint arXiv:2509.24291},
  year={2025}
}

@article{AMER,
  title={Beyond Single Embeddings: Capturing Diverse Targets with Multi-Query Retrieval},
  author={Chen, Hung-Ting and Liu, Xiang and Ravfogel, Shauli and Choi, Eunsol},
  journal={arXiv preprint arXiv:2511.02770},
  year={2025}
}

@article{AdaQR,
  title={Your Dense Retriever is Secretly an Expeditious Reasoner},
  author={Zhang, Yichi and Bai, Jun and Cai, Zhixin and Qin, Shuhan and Chen, Zhuofan and Guan, Jinghua and Rong, Wenge},
  journal={arXiv preprint arXiv:2510.21727},
  year={2025}
}

@inproceedings{laser,
author = {Jin, Jiajie and Zhang, Yanzhao and Li, Mingxin and Long, Dingkun and Xie, Pengjun and Zhu, Yutao and Dou, Zhicheng},
title = {Internalizing Explicit Reasoning into Latent Space for Dense Retrieval},
year = {2026},
isbn = {9798400725999},
publisher = {Association for Computing Machinery},
address = {New York, NY, USA},
url = {https://doi.org/10.1145/3805712.3809575},
doi = {10.1145/3805712.3809575},
booktitle = {Proceedings of the 49th International ACM SIGIR Conference on Research and Development in Information Retrieval},
pages = {689--700},
numpages = {12},
location = {Australia},
series = {SIGIR '26}
}

@article{xetrieval,
  title={Xetrieval: Mechanistically Explaining Dense Retrieval},
  author={Cai, Zhixin and Bai, Jun and Liu, Yang and Li, Jiaqi and Zhang, Yichi and Li, Taichuan and Chen, Zhuofan and Jia, Zixia and Zheng, Zilong and Rong, Wenge},
  journal={arXiv preprint arXiv:2605.29507},
  year={2026}
}

\clearpage
\appendix
\section*{Supplementary Material}
\subsection*{Overview}
This supplement provides the implementation details needed to reproduce
\method{} without repeating the methodology in the main paper. We focus on
three core aspects: \textit{how the CoT sequences are structured and segmented},
\textit{how training is made tractable under the memory pressure imposed by
ultra-long CoT branches and a cross-device document pool, chiefly through a
deterministic gradient replay strategy}, and \textit{how the evaluation protocols
and efficiency optimizations (including KV-cache reuse) are implemented}.
Explicit CoT is used exclusively during training. At inference, a query is
represented by the final silent state and each document remains a reusable
single vector.

\section{Structured Segmentation of CoT}
\label{sec:supp_data}

The CoT sequences in this dataset typically comprise four stages: problem identification, reasoning and relevant information, detailed solution, and verification and answer convergence. We introduce a robust parsing mechanism to systematically segment the continuous text into these four distinct stages.
To overcome the lack of standardized paragraph boundaries in raw texts, our parser adopts a two-tier strategy. First, it prioritizes high-level semantic headings and structural step markers while carefully avoiding the misclassification of ordinary numbered calculations or code lines. Second, when encountering highly non-standard texts with insufficient reliable boundaries, the parser gracefully degrades by identifying natural line breaks or sentence boundaries near the four equal-length partitions of the text. This ensures reliable and structured supervision without disrupting the logical coherence of the chain of thought.

Furthermore, to prevent this segmentation operation from disrupting the original logical context, we do not let the model learn these four segmented slices $(r_1, r_2, r_3, r_4)$ in isolation. Instead, we formulate them as cumulative targets:
\begin{equation}
  r_{\leq k}=r_1\Vert r_2\Vert\cdots\Vert r_k,
  \qquad k\in\{1,2,3,4\},
\label{eq:supp_cumulative}
\end{equation}
where $\Vert$ denotes concatenation with a blank line. Through this cumulative mechanism, the original four continuous slices are reorganized so that each subsequent reasoning stage naturally retains and observes the prior deduction history. This perfectly maintains the logical coherence of the chain of thought while enabling stage-wise supervision.

\section{Training under Memory Constraints}
\label{sec:supp_model}
Table~\ref{tab:supp_training} details our training configuration, which relies on a frozen Qwen3-8B base augmented with LoRA (rank 32) and RGLT (rank 8) parameters. Training is based on a joint objective that combines four terms: final retrieval, per-stage retrieval credit, CoT reconstruction, and a retrieval-effect match between silent-stage transitions and the explicit CoT-prefix transition, while reusing the same frozen backbone, LoRA, and RGLT parameters throughout. Concretely, the forward passes that must coexist in memory on every step include the explicit CoT branches (up to 8,192 tokens) for the sampled cumulative prefixes, the document embeddings (one positive plus three hard negatives per query, shared across the batch), and the before/after pairs fed to the effect and reconstruction objectives, all multiplied by the cross-device document pool of 512. Retaining the computation graph for all of these simultaneously is infeasible even on $4\times8$ A100 80GB GPUs. We therefore adopt deterministic gradient replay. The representations for these branches are first computed in micro-batches and detached as leaf tensors. Once the loss backward pass yields the representation gradients, the corresponding forwards are replayed to backpropagate those exact gradients into the weights. To guarantee mathematical equivalence between the initial forward pass and the replay, we enforce strict zero dropout across all model layers and LoRA modules. The much shorter 512-token latent query graph remains attached normally.

\begin{table}[t]
\centering
\small
\caption{Training configuration.}
\label{tab:supp_training}
\setlength{\tabcolsep}{4.6pt}
\begin{tabular}{ll}
\toprule
Setting & Value \\
\midrule
Backbone & Qwen3-8B \\
GPU & $4\times8$ NVIDIA A100 80GB \\
Precision / attention & bfloat16 / FlashAttention 2 \\
Per-GPU / global query batch & 4 / 128 \\
Per-query candidates & 1 positive + 3 hard negatives \\
Cross-device document pool & 512 \\
Query/document maximum length & 512 tokens \\
CoT maximum length & 8,192 tokens, no truncation \\
Silent tokens / stages & 16 / 4 \\
LoRA rank / scale / dropout & 32 / 64 / 0 \\
RGLT rank & 8 \\
LoRA / mechanism learning rate & $10^{-5}$ / $5\times10^{-5}$ \\
Optimizer / weight decay & AdamW / 0.01 \\
Warm-up / schedule & 100 steps / cosine decay \\
Gradient clipping & 1.0 \\
\bottomrule
\end{tabular}
\end{table}

\section{Evaluation Details and Inference Optimization}
\label{sec:supp_evaluation}

\subsection{Benchmark Protocol}
\label{sec:supp_protocol}

\paragraph{BRIGHT.}
We evaluate all 1,384 queries and complete corpora from the 12 domains.  Query
instructions are enabled, and the maximum length is 512. LeetCode is split into four
document shards and merged before scoring.  The full evaluation recomputes
document vectors, and cache metadata contains the model and checkpoint
fingerprint.  We report the unweighted macro average of domain nDCG@10 and
Recall@10.

\paragraph{FollowIR.}
We use the official MTEB 1.38.32 Robust04, News21, and Core17 instruction
retrieval tasks.  Their standard metrics are MAP@5, nDCG@5, and MAP@5,
respectively.  The query and instruction are passed through the separate
channels used by the contextual anchor; documents remain instruction
independent.  In addition to the standard task metrics, we use the official
p-MRR implementation to measure whether changed instructions move newly
relevant documents upward and newly irrelevant documents downward.  We macro-average each metric over the three tasks.

\paragraph{BrowseComp-Plus.}
We use the benchmark's fixed corpus, queries, and relevance labels and report
Recall@5, Recall@100, and Recall@1000.  The document index is built once from
the document-only path and is not modified at query time.

\subsection{KV-Cache Inference and Efficiency Measurement}
\label{sec:supp_efficiency}

\textbf{Query KV-Cache Reuse.} 
Naively running the anchor and latent paths as two complete forward passes would encode the query twice, roughly doubling the latency. To solve this, our implementation strictly reuses the causal query-prefix key/value cache. Let $P(q)$ be the complete tokenized prefix shared by both paths, ending immediately after the query. Inference is decomposed as follows:
\begin{equation}
\begin{aligned}
  C_q &= \operatorname{Prefill}(P(q)),\\
  a   &= \operatorname{Forward}(I;\anchor\mid C_q)_{\anchor},\\
  z_4 &= \operatorname{Forward}\bigl(A(a);\\
     &\qquad\quad \emb;T_{1:K}\mid C_q\bigr)_{T_{K}},
\end{aligned}
\label{eq:supp_kv_cache}
\end{equation}
where $C_q$ contains the per-layer keys and values of the shared prefix. The anchor branch and latent branch receive separate cache objects that map to the same read-only prefix tensors. Crucially, to prevent memory bloat, the instruction and temporary anchor cache are immediately discarded after $a$ is obtained. The latent branch then spawns directly from the original $C_q$, completely bypassing the instruction text.

\textbf{Preserving Positional Integrity.} 
When sharing caches across variable-length batches, misaligned position IDs can silently corrupt the representations. To guarantee mathematical equivalence to full recomputation, we enforce strict positional tracking. Prefixes are left-padded, and suffix position IDs are dynamically computed per row starting exactly from each example's non-padding prefix length. Furthermore, to maximize GPU utilization, all $K$ silent tokens are processed concurrently in a single causal block rather than being decoded autoregressively token-by-token. 

\textbf{Computational Overhead Analysis.} 
Because of the above designs, the query prefix is encoded only once. The additional computational burden relative to a standard base retriever is strictly bounded to two short suffix blocks: the instruction-plus-anchor suffix and the fixed ($K+2$)-token latent block (injected anchor, one readout, $K$ silent tokens). This architectural isolation is the precise reason why our measured latency overhead is only ${\sim}1.2\times$, rather than the $2\times$ cost of two full forwards.

\enlargethispage{2.5\baselineskip}
\textbf{Strict Validation and Profiling.} 
To ensure our optimization does not compromise retrieval accuracy, we enforce a strict cache-correctness test. Cached inference must yield identical top-$k$ rankings compared to full uncached recomputation, with the maximum absolute difference of ST(final) embeddings bounded near floating-point epsilon. For latency benchmarking, we isolate the components by using \texttt{torch.cuda.synchronize()} around the prefill, anchor suffix, and latent suffix regions separately, explicitly excluding I/O-bound operations like tokenization and index construction.

\end{document}